%% file: santesteban_EG2020.tex
\documentclass{egpubl}
\usepackage{eg2020}
 
\ConferencePaper        %

\usepackage[T1]{fontenc}
\usepackage{dfadobe}  

\usepackage{cite}  %

\BibtexOrBiblatex
\electronicVersion
\PrintedOrElectronic
\ifpdf \usepackage[pdftex]{graphicx} \pdfcompresslevel=9
\else \usepackage[dvips]{graphicx} \fi

\usepackage{egweblnk}

\usepackage{scalerel}
\usepackage{amsfonts}
\usepackage{amsmath}
\usepackage{physics}
\usepackage[dvipsnames]{xcolor}
\usepackage{booktabs}
\usepackage[normalem]{ulem}

\newcommand{\etal}{et al.}
\newcommand{\eg}{e.g.}
\newcommand{\ie}{i.e.}

\newcommand{\revised}[1]{#1}
\newcommand{\removed}[1]{}

\title[SoftSMPL: Data-driven Modeling of Nonlinear Soft-tissue Dynamics for Parametric Humans]%
{SoftSMPL: Data-driven Modeling of Nonlinear Soft-tissue Dynamics for Parametric Humans}

\author[I. Santesteban, E. Garces, M.A. Otaduy \& D. Casas]
{\parbox{\textwidth}{\centering 
        Igor Santesteban~~~~~~~~~~~~~~   
		Elena Garces~~~~~~~~~~~~~~
		Miguel A. Otaduy~~~~~~~~~~~~~~
        Dan Casas
}
         \\
{\parbox{\textwidth}{\centering Universidad Rey Juan Carlos, Madrid, Spain}
}
}

\begin{document}
	
	\teaser{
	 \includegraphics[width=0.8\linewidth]{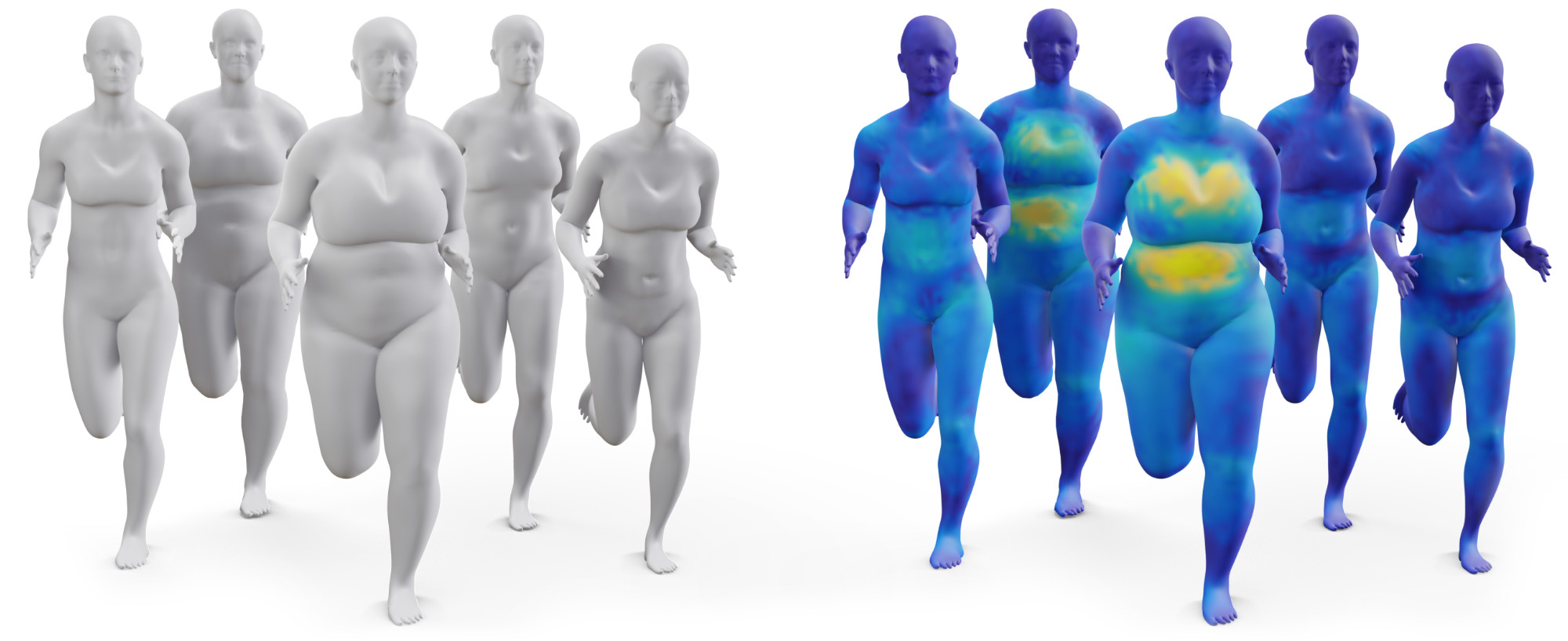}
	 \centering
	  \caption{Our method regresses soft-tissue dynamics for parametric avatars. Here we see five different body shapes performing a running motion, each of them enriched with soft-tissue dynamics. We depict the magnitude of the regressed displacements using colormaps (right). See supplementary video for full animation.}
	\label{fig:teaser}
	}
	
	\maketitle
	\begin{abstract}
	We present SoftSMPL, a learning-based method to model realistic soft-tissue dynamics as a function of body shape and motion. Datasets to learn such task are scarce and expensive to generate, which makes training models prone to overfitting. At the core of our method there are three key contributions that enable us to model highly realistic dynamics and better generalization capabilities than state-of-the\revised{-}art methods, while training on the same data. 
	First, a novel motion descriptor that disentangles the standard pose representation by removing subject-specific features; 
	second, a neural-network-based recurrent regressor that generalizes to unseen shapes and motions; 
	and third, a highly efficient nonlinear deformation subspace capable of representing soft-tissue deformations of arbitrary shapes. 
	We demonstrate qualitative and quantitative improvements over existing methods and, additionally, we show the robustness of our method on a variety of motion capture databases. 
		
		

	\end{abstract}  

\input{intro.tex}

\input{related-work.tex}

\input{overview.tex}

\input{model.tex}

\input{disentanglement.tex}

\input{data.tex}

\input{evaluation.tex}
\input{conclusions.tex}

\bibliographystyle{eg-alpha-doi} 
\bibliography{santesteban_EG2020}   
\end{document}

%% file: intro.tex
\section{Introduction} \label{sec:intro}
Soft-tissue dynamics are fundamental to produce compelling human animations. 
Most of existing methods capable \removed{to generate}\revised{of generating} highly dynamic soft-tissue deformations are based on  physics-based approaches. However, these methods are challenging to implement due to the inner complexity of the human body, and the expensive simulation process needed to animate the model.
Alternatively, data-driven models can potentially learn human soft-tissue deformations as a function of body pose directly from real-world data (\textit{\eg}, 3D reconstructed sequences). However, in practice, this is a very challenging task due to the highly nonlinear nature of the dynamic deformations, and the scarce of datasets with sufficient reconstruction fidelity.

In this work we propose a novel learning-based method to animate parametric human models with highly expressive soft-tissue dynamics. 
SoftSMPL\removed{Our animation model} takes as input the shape descriptor of a body and a motion descriptor, and produces dynamic soft-tissue deformations that generalize to unseen shapes and motions.
Key to our method is to realize that humans move in a highly personalized manner, \textit{i.e.}, motions are shape and subject dependent, and \removed{this} \revised{these} subject-dependant features are usually entangled in the pose representation.

Previous methods fail to disentangle body pose from shape and subject features;
therefore, they overfit the relationship between tissue deformation and pose, and generalize poorly to unseen shape\revised{s} and motions. 
Our method overcomes this limitation by proposing a new  representation to disentangle the traditional pose space in two steps.
First, we propose a solution to encode a compact and {\em deshaped} representation of body pose which eliminates the correlation between individual \textit{static} poses and subject.
Second, we propose a motion transfer approach, which uses person-specific models to synthesize animations for pose (and style) sequences of other persons. 
As a result, our model is trained with data where pose and subject-specific \textit{dynamic} features are no longer entangled. 
We complement this contribution with a highly efficient nonlinear subspace to encode tissue deformations of arbitrary bodies, and a neural-network-based recurrent regressor as our learning-based animation model. 
We demonstrate qualitative and quantitative improvements over previous methods, as well as robust performance on a variety of motion capture databases.

%% file: related-work.tex
\section{Related work}
The 3D modeling of human bodies has been investigated following two main trends: data-driven models, which learn deformations directly from data; and physically-based models, which compute body deformations by solving a simulation problem, usually consisting of a kinematic model coupled with a deformable layer. In this section we discuss both trends, with special emphasis on the former, to which our method belongs.   
 
\subparagraph*{Data-driven models.}Pioneering data-driven models interpolate manually sculpted static 3D meshes to generate new samples 
\cite{sloan2001shapeexample}. 
With the appearance of laser scanning technologies, capable of reconstructing 3D static bodies with great level of detail, the data-driven field became popular. 
Hilton \textit{et al.} \cite{hilton2002animatedmodel} automatically fit an skeleton to a static scan to generate animated characters.
Allen \textit{et al.} proposed one of the first methods to model upper body 
\cite{allen2002articulated} and full body
\cite{allen2003humanbody} deformations using a shape space learned from static scans and an articulated template.
Anguelov \textit{et al.} \cite{anguelov2005scape} went one step further and modeled both shape \textit{and} pose dependent deformations directly from data. 
Many follow-up data-driven methods have appeared \cite{hasler2009statistical,jain2010moviereshape,hirshberg2012coregistration,chen2013tensor,yang2014semantic,feng2015reshaping,zuffi2015stitched,loper_SIGAsia2015,pishchulin2017building}, but all of these are limited to modeling \textit{static} deformations.

Data-driven models have also been explored to model soft-tissue deformations, which is our main goal too.
Initial works used \textit{sparse} marker-based systems to acquire the data.
The pioneering work of Park and Hodgins \cite{park2006capturing} reconstructs soft-tissue motion of an actor by fitting a 3D mesh to 350 tracked points.
In subsequent work \cite{park2008datadriven}, they proposed a second-order dynamics model to synthesize skin deformation as a function of body motion. Similar to us, they represented both body pose and dynamic displacements in a low-dimensional space. However, their method does not generalize to different body shapes.
Neumann \textit{et al.} \cite{neumann2013capture} also used sparse markers to capture shoulder and arm deformations of multiple subjects in a multi-camera studio. They were able to model muscle deformations as a function of shape, pose, and external forces, but their method is limited to the shoulder-arm area, and cannot learn temporal dynamics. Similarly, Loper \textit{et al.} \cite{loper2014mosh} did not learn dynamics either, but they were able to estimate full body pose \textit{and} shape from a small set of motion capture markers. Remarkably, despite their lack of explicit dynamics, their model can reproduce soft-tissue motions by allowing body shape parameters to change over time. 

More recently, 3D/4D scanning technologies and mesh registration methods~\cite{bradley2008garmentcapture,cagniart2010probabilistic,dou20153d,bogo2017dfaust,robertini2017surfacedetails,ponsmoll2017clothcap} allow to reconstruct high-quality dynamic sequences of human performances.
These techniques have paved the way to data-driven methods that leverage \textit{dense} 3D data, usually in the form of temporally coherent 3D mesh sequences, to extract deformation models of 3D humans. 
Neumann \textit{\etal} \cite{neumann2013splocs} used 3D mesh sequences to learn sparse localized deformation modes, but did not model temporal dynamics.
Tsoli \textit{\etal} reconstructed 3D meshes of people breathing  \cite{tsoli2014breathing} in different modes, and built a statistical model of body surface deformations as a function of lung volume. In contrast, we are able to model far more complex deformations, with higher frequency dynamics, as a function of body shape and pose.
Casas and Otaduy~\cite{casas_PACMCGIT2018} modeled full-body soft-tissue deformations as a function of body motion using a neural-network-based nonlinear regressor. Their model computes per-vertex 3D offsets encoded in an efficient subspace, however it is subject-specific and does not generalize to different body shapes.
Closest to our work is Dyna \cite{ponsmoll2015dyna}, a state-of-the-art method that relates soft-tissue deformations to motion and body shape from 4D scans. Dyna uses a second-order auto-regressive model to output mesh deformations encoded in a subspace. Despite its success in modeling surface dynamics, we found that its generalization capabilities to unseen shapes and poses are limited due to the inability to effectively disentangle pose from shape and subject-style \removed{deformations}.
Furthermore, Dyna relies on a linear PCA subspace to represent soft-tissue deformations, which struggles to reproduce highly non-linear deformations.
DMPL~\cite{loper_SIGAsia2015} proposes a  soft-tissue deformation model heavily-inspired in Dyna, with the main difference that it uses a vertex-based representation instead of triangle-based. However, DMPL suffers from the same limitations of Dyna mentioned above. 
We also propose a vertex-based representation, which eases the implementation in standard character rigging pipelines, while achieving superior generalization capabilities and more realistic dynamics.

Garment and clothing animation have also been addressed with data-driven models that learn surface deformations as a function of human body parameters \revised{\cite{deAguiar2010, wang2019garmentAuthoring}}.
Guan \textit{\etal}~\cite{guan2012drape}, \removed{use a linear model to regress wrinkles from pose parameters, but do not learn shape-dependent deformations}
\revised{use linear models to regress wrinkles from pose and shape parameters, but each source of deformation is modeled independently.} Similarly, other data-driven methods are limited to represent garment shape variations as linear scaling factors \cite{ponsmoll2017clothcap,yang2018analyzing, laehner2018deepwrinkles} and therefore do not reproduce realistic deformations.
Closer to ours is the work of Santesteban \textit{\etal}~\cite{santesteban_EG2019} which effectively disentangles garment deformations due to shape and pose, allowing to train a \revised{nonlinear} deformation regressor that generalizes to new subjects and motions.
Alternatively, Gundogdu \textit{\etal}~\cite{gundogdu2019garnet}, extracts geometric features of human bodies to use them to parameterize garment deformations.

\subparagraph*{Physically-based models.}
The inherent limitation of data-driven models is their struggle to generate deformations far from the training examples.
Physically-based models overcome this limitation by formulating the deformation process within a simulation framework. However, these approaches are not free of difficulties: defining an accurate and efficient mechanical model to represent human motions, and solving the associated simulations is hard.

Initial works used layered representations consisting of a deformable volume for the tissue layer,
rigidly attached to a kinematic skeleton~\cite{capell2002dyanmicdeform,larboulette2005dynamic}.
Liu \textit{\etal}~\cite{liu2013softbody} coupled rigid skeletons for motion control with a pose-based plasticity model to enable two-way interaction between skeleton, skin, and environment.
McAdams \textit{\etal}~\cite{mcadams2011efficientelasticity} showed skin deformations with a discretization of corotational elasticity on a hexahedral lattice around the surface mesh, but did not run at real-time rates.
\revised{Xu and Barbi\v{c}~\cite{xu2016} used secondary Finite Element Method (FEM) dynamics and model reduction techniques to efficiently enrich the deformation of a rigged character.}
To speed up simulations, Position-Based Dynamics (PBD)~\cite{bender2017survey} solvers have been widely used for different physics systems, also for human soft tissue~\cite{bender2013physicsskinning,komaritzan2018projective} and muscle deformation~\cite{romeo2019muscle}. 
Projective Dynamics, another common approach to accelerate simulations, has also been used for simulating deformable characters~\cite{li2019fastsimulation}. 
\revised{Meanwhile, Pai \textit{\etal}~\cite{pai2018humantouch} presented a novel hand-held device to estimate the mechanical properties of real human soft-tissue.}

\paragraph*{Subspaces for Simulation.}
Subspace simulation methods attempt to find a low-dimensional representation of an initial dense set of equations in order to facilitate the computations. 
PCA has been widely used to this end \cite{krysl2001dimensional,barbivc2005real,treuille2006model}, while an alternative set of works aims to find new and more efficient bases \cite{teng2014simulating, teng2015subspace}.
For clothing, Hahn \textit{\etal}~\cite{hahn2014subspace} built a linear subspace using temporal bases distributed across pose space.
Holden \textit{\etal}~\cite{holden2019subspacephysics} also built a linear subspace using PCA and machine learning to model external forces and collisions at interactive rates. 
Finally, Fulton \textit{\etal}~\cite{Fulton:LSD:2018} built a non-linear subspace using an auto-encoder on top of an initial PCA subspace to accelerate the solver.

%% file: overview.tex
\begin{figure*}
	\includegraphics[width=\linewidth]{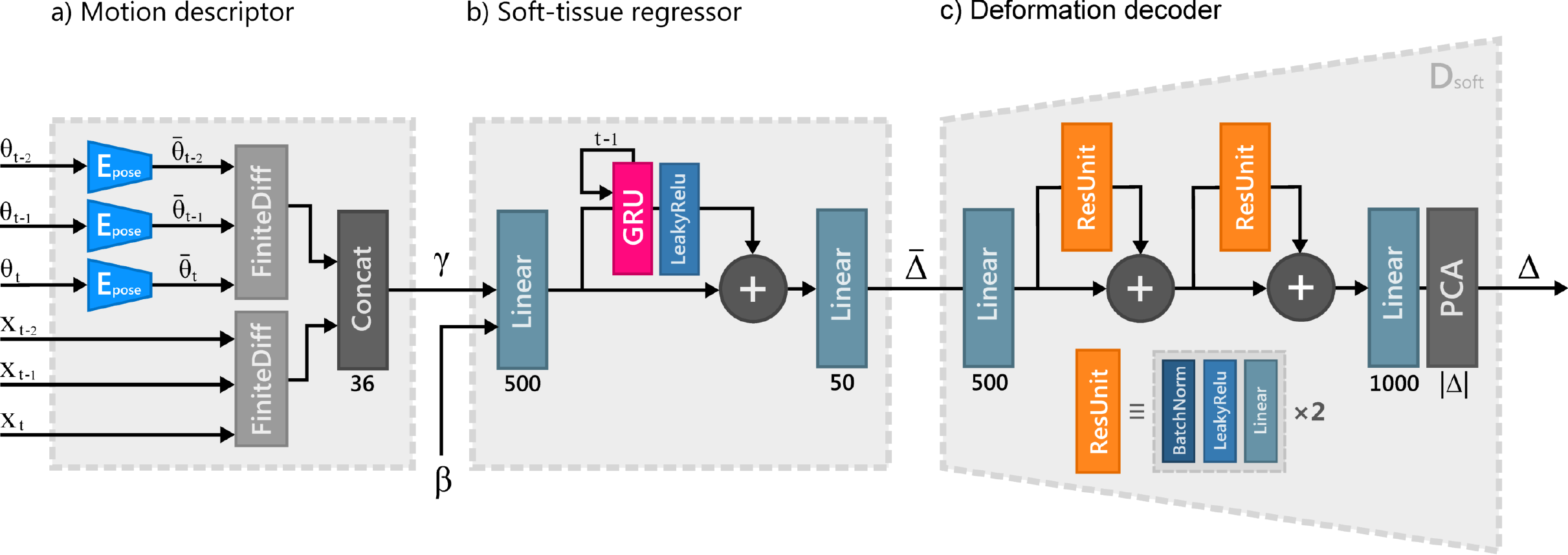}
	\caption{Runtime pipeline of our approach. First, the temporal motion data is encoded in our novel disentangled pose descriptor. Then, the resulting low dimensional vector is concatenated with the skeleton root offsets to form the motion descriptor. This descriptor along with the desired shape parameters are passed through the soft-tissue regressor, which predicts the nonlinear dynamic behaviour of the soft-tissue deformation in a latent space. Finally, the deformation decoder recovers the original full space of deformation offsets for each vertex of the mesh.}
	\label{fig:pipeline}
\end{figure*}
\section{Overview}
Our animation model for soft-tissue dynamics takes as input descriptors of body shape and motion, and outputs surface deformations.
These deformations are represented as per-vertex 3D displacements of a human body model, described in Section \ref{sec:softtissue_model}, and encoded in an efficient nonlinear subspace, described in Section \ref{sec:soft_tissue_subspace}.
At runtime, given body and motion descriptors, we predict the soft-tissue deformations using a novel recurrent regressor proposed in Section \ref{sec:regressor}. 
Figure \ref{fig:pipeline} depicts the architecture of our runtime pipeline, including the motion descriptor, the regressor, and a soft-tissue decoder to generate the predicted deformations.

In addition to our novel subspace and regressor, our key observation to achieve highly expressive dynamics with unprecedented generalization capabilities is an effective disentanglement of the pose space.
In Section \ref{sec:disentanglement}, we argue and demonstrate that the standard pose space (\textit{i.e.}, vector of joint angles $\theta$) used in previous methods is entangled with subject-specific features.
This causes learning-based methods to overfit the relationship between tissue deformation and pose.
In Section \ref{sec:motion_descriptor} we identify \textit{static} features, mostly due to the particular anatomy of each person, that are entangled in the pose space, and propose a \textit{deshaped} representation to effectively disentangle them. Furthermore, in Section \ref{sec:motion_transfer} we identify \textit{dynamic} features that manifest across a sequence of poses (also known as \textit{style}), and propose a strategy to eliminate them.

%% file: model.tex
\section{SoftSMPL}
\label{sec:data-driven-dynamics}
\begin{figure}
    \centering
    \includegraphics[width=\columnwidth]{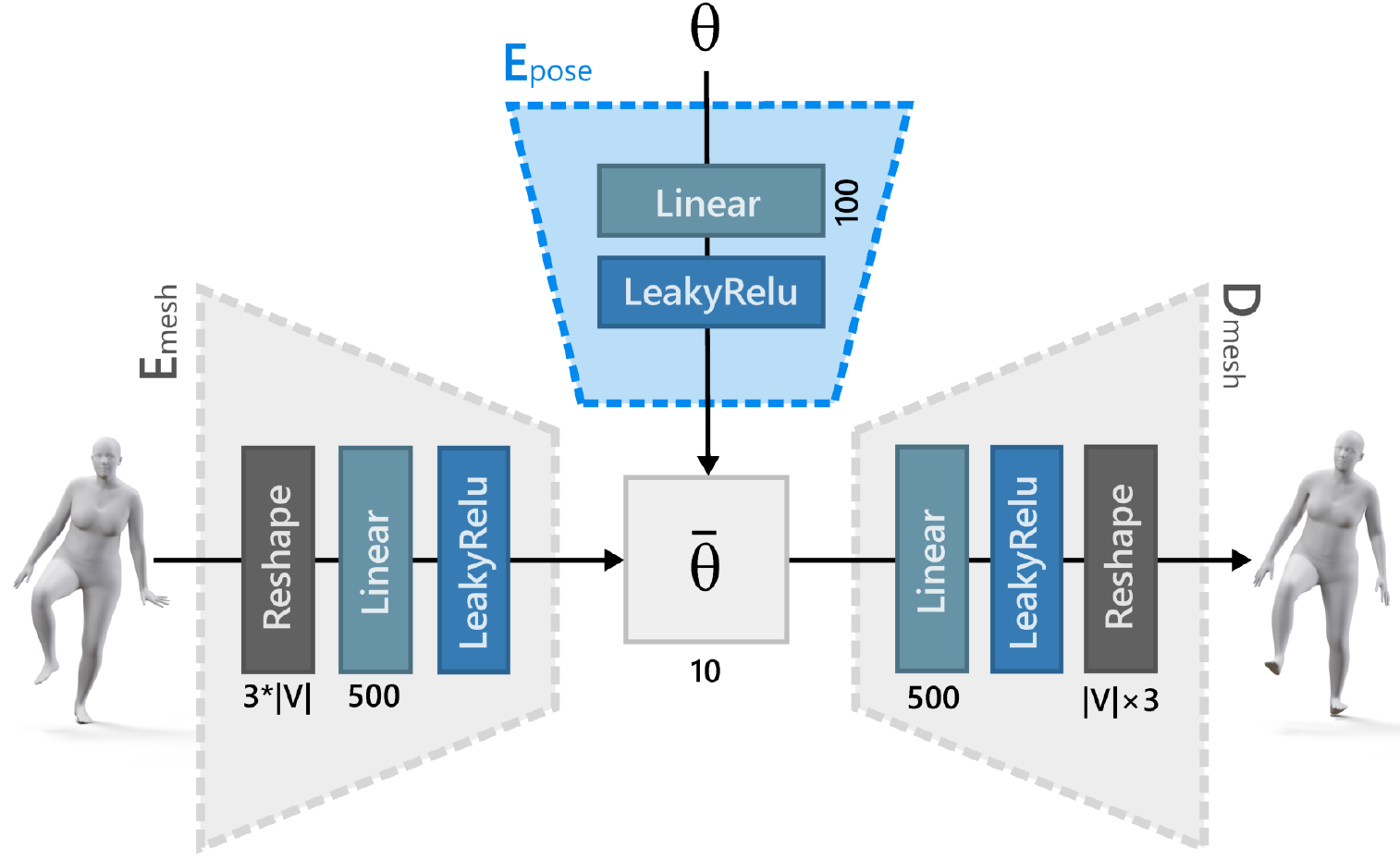}
    \caption{Architecture of the multi-modal pose autoencoder.}
    \label{fig:pose_descriptor}
\end{figure}
\subsection{Human Model}
\label{sec:softtissue_model}
We build our soft-tissue model on
top of standard human body models~ (\textit{\eg},~\cite{feng2015reshaping,loper_SIGAsia2015}) controlled by shape parameters $\beta \in \mathbb{R}^{|\beta|}$ (\textit{\eg}, principal components of a collection of body scans in rest pose) and pose parameters $\theta  \in \mathbb{R}^{|\theta|}$(\textit{\eg}, joint angles).
These works assume that a deformed body mesh $M(\beta, \theta)  \in \mathbb{R}^{3\times V}$, where $V$ is the number of vertices, is obtained by
\begin{equation}
M(\beta, \theta) = W(T(\beta,\theta),\beta, \theta,\mathcal{W}) 
\end{equation}
where $W(\cdot)$ is a skinning function (\textit{e.g.}, linear blend skinning, dual quaternion, etc.) with skinning weights $\mathcal{W}$ that deforms an unposed body mesh $T(\beta,\theta)  \in \mathbb{R}^{3\times V}$.

Inspired by Loper \textit{\etal}~\cite{loper_SIGAsia2015},  who obtain  the unposed mesh $T(\beta,\theta)$ by
deforming a body mesh template  $\mathbf{T}\in \mathbb{R}^{3\times V}$ to incorporate changes in shape $B_\text{s}(\beta)$ and pose corrective displacements $B_\text{p}(\theta)$, we propose to further deform the body mesh template to incorporate soft-tissue dynamics.
More specifically, we define our unposed body mesh as
\begin{equation}
T(\beta,\theta,\gamma) = \mathbf{T} + B_\text{s}(\beta) + B_\text{p}(\theta) + 
B_\text{d}(\gamma,\beta),
\label{eq:deformed_template}
\end{equation}
where $B_\text{d}(\gamma,\beta) = \Delta \in\mathbb{R}^{3 \times V}$ 
is a soft-tissue regressor that outputs per-vertex displacements required to reproduce skin dynamics given a shape parameter $\beta$ and a motion descriptor $\gamma$.
Notice that, in contrast to previous model-based works that also predict soft-tissue displacements \cite{ponsmoll2015dyna,loper_SIGAsia2015,casas_PACMCGIT2018}, our key observation is that such regressing task cannot be formulated directly as function of pose $\theta$ (and shape $\beta$), because subject-specific information is entangled in that pose space. See Section \ref{sec:disentanglement} for a detailed description of our motion descriptor $\gamma$ and full details on our novel pose disentanglement method. 
\subsection{Soft-Tissue Representation and Deformation Subspace}
\label{sec:soft_tissue_subspace}
We represent soft-tissue deformations $\Delta$ as per-vertex 3D offsets of a body mesh $\mathbf{T}$ in an unposed state. 
This representation\removed{s} allows to isolate the soft-tissue deformation component from other deformations, such as pose or shape.

\pagebreak
Given the data-driven nature of our approach, in order to train our model it is crucial that we define a strategy to extract ground truth deformations $\Delta^{\scaleto{\text{GT}}{4pt}} \in\mathbb{R}^{3 \times V}$ from real world data. 
To this end, \revised{in a similar spirit to \cite{ponsmoll2015dyna, loper_SIGAsia2015, ponsmoll2017clothcap},} given a dataset $\mathcal{S} = \{\textbf{S}_t\}_{t=0}^{T-1}$ of 4D scans with temporally consistent topology, we extract the soft-tissue component of each mesh $\textbf{S} \in\mathbb{R}^{3 \times V}$ as 
\begin{equation}
\Delta^{\scaleto{\text{GT}}{4pt}} = W^{-1}(\mathbf{S}, \theta, \mathcal{W}) - \mathbf{T} - B_\text{P}(\theta) - B_\text{S}(\beta),
\label{eq:displacement_gt}
\end{equation}
where $W^{-1}(\cdot)$ is the inverse of the skinning function, $B_\text{P}(\theta)$ a corrective pose blendshape, and  $B_\text{S}(\beta)$ a shape deformation blendshape (see \cite{loper_SIGAsia2015} for details on how the latter two are computed). 
Solving Equation \ref{eq:displacement_gt} requires estimating the pose $\theta$ and shape $\beta$ parameters for each mesh $\mathbf{S}$, which is a priori unknown (\textit{\ie}, the dataset $\mathcal{S}$ contains only 3D meshes, no shape or pose parameters).
Similar to \cite{ponsmoll2015dyna}, we solve the optimization problem:
\begin{equation}
\operatorname*{argmin}_{\theta,\beta} ||\mathbf{S}-M(\theta,\beta)||_2
\label{eq:fitting}
\end{equation}
to estimate the shape $\beta$ and pose $\theta$ parameters of each scan $\mathbf{S}$ in the dataset $\mathcal{S}$.

Despite the highly-convenient representation of encoding soft-tissue deformations as per-vertex 3D offsets $\Delta\in\mathbb{R}^{3 \times V}$, this results in a too high-dimensional space for an efficient learning-based framework. Previous works \cite{loper_SIGAsia2015,ponsmoll2015dyna} use linear dimensionality reduction techniques (\textit{\eg}, Principal Component Analysis) to find a subspace capable of reproducing the deformations without significant loss of detail. However, soft-tissue deformations are highly nonlinear, hindering the reconstructing capabilities of linear methods. 
We mitigate this by proposing a novel autoencoder to find an efficient nonlinear subspace to encode soft-tissue deformations of parametric humans.

Following the standard autoencoder pipeline, we define the reconstructed (\textit{\ie},  encoded-decoded) soft-tissue deformation as
\begin{equation}
\Delta_\text{rec} = D_\text{soft}(E_\text{soft}(\Delta)),
\end{equation}
where $\bar{\Delta} = E_{\text{soft}}(\Delta)$ and $D_{\text{soft}}(\bar{\Delta})$ are  encoder and decoder networks, respectively, and $\bar{\Delta} \in \mathbb{R}^{|\bar{\Delta|}}$ soft-tissue displacements projected into the latent space.
We train our deformation autoencoder by using a loss function $\mathcal{L}_{\text{rec}}$ that minimizes both surface and normal errors between input and output displacements as follows
\begin{eqnarray}
\mathcal{L}_\text{surf}&=& \norm{\Delta -\Delta_\text{rec}}_2
\\
\mathcal{L}_\text{norm}&=& \frac{1}{F}\sum_{f=1}^F\norm{1 - N_f(\Delta)\cdot N_f(\Delta_{\text{rec}})}_1
\\
\mathcal{L}_{\text{rec}} &=& \mathcal{L}_\text{surf} +  \lambda_\text{norm}\mathcal{L}_\text{norm}
\end{eqnarray}
where $F$ is the number of faces of the mesh template, $N_f(\Delta)$ the normal of the $f^{\text{th}}$ face, and $\lambda_\text{norm}$ is set to 1000.
Notice that, during training, we use ground truth displacements $\Delta^{\scaleto{\text{GT}}{4pt}}$ from a variety of characters which enables us to find a subspace that generalizes well to encode soft-tissue displacements of \textit{any} human shape.
This is in contrast to previous works \cite{casas_PACMCGIT2018} that need to train shape-specific autoencoders.

We implement the encoder $E_{\text{soft}}$ and decoder $D_{\text{soft}}$ using a \revised{fully-connected} neural network architecture \revised{composed of several residual units \cite{he2016identity}} \removed{that combines linear and residual layers}. \revised{Inspired by the work of Fulton \textit{\etal}~\cite{Fulton:LSD:2018}, we initialize the first and last layers of the autoencoder with weights computed using PCA, which eases the training of the network.} In Figure 	\ref{fig:pipeline} (right) we depict the decoder $D_{\text{soft}}$. The encoder $E_{\text{soft}}$ uses an analogous architecture.

\subsection{Soft-Tissue Recurrent Regressor}
\label{sec:regressor}
In this section we describe the main component of our runtime pipeline: the soft-tissue regressor $R$, illustrated in Figure \ref{fig:pipeline} (center).
Assuming a motion descriptor $\gamma$ (which we discuss in detail in Section \ref{sec:motion_descriptor}) and a shape descriptor $\beta$, our regressor outputs the predicted soft tissue displacements $\bar{\Delta}$. These encoded displacements are subsequently fed into the decoder $D_\text{soft}$ to generate the final per-vertex 3D displacements 
\begin{equation}
\Delta = D_{\text{soft}}(R(\gamma, \beta)).
\end{equation}

To learn the naturally nonlinear dynamic behavior of soft-tissue deformations, we implement the regressor $R$ using a recurrent architecture GRU \cite{cho2014learning}.
Recurrent architectures learn which information of previous frames is relevant and which not, resulting in a good approximation of the temporal dynamics.
This is in contrast to modeling temporal dependencies by explicitly adding the output of one step as the input of the next step, which is prone to instabilities specially in nonlinear models. 
Furthermore, our regressor also uses a residual shortcut connection to skip the GRU layer altogether, which improves the flow of information \revised{\cite{he2016residual}}. \revised{We initialize the state of the GRU to zero at the beginning of each sequence.}

We train the regressor $R$ by minimizing a loss $\mathcal{L}_{\text{reg}}$, which enforces predicted vertex positions, velocities, and accelerations to match the latent space deformations $\bar{\Delta}$, 
\begin{equation}
\mathcal{L}_{\text{reg}} = \mathcal{L}_{\text{pos}} +  \mathcal{L}_{\text{vel}} + \mathcal{L}_{\text{acc}}
\end{equation}

%% file: disentanglement.tex
\section{Disentangled Motion Descriptor}
\label{sec:disentanglement}
To efficiently train the soft-tissue regressor  $R(\gamma,\beta)$, described earlier in Section \ref{sec:regressor}, we require a pose-disentangled and discriminative motion descriptor $\gamma$.
To this end, in this section we propose a novel motion descriptor. It encompasses the velocity and acceleration of the body root in world space $X$, a novel pose descriptor $\bar{\theta}$, and the velocity and acceleration of this novel pose descriptor, as follows:
\begin{equation}
\gamma = \{\bar{\theta}, \dv{\bar{\theta}}{t},\dv[2]{\bar{\theta}}{t}, \dv{X}{t}, \dv[2]{X}{t}\}.
\end{equation}

In the rest of this section we discuss the limitation of the pose descriptors used in state-of-the-art human models, and introduce a new disentangled space $\bar{\theta}$ to remove \textit{static} subject-specific features (Section \ref{sec:motion_descriptor}). Moreover, we also propose a strategy to remove \textit{dynamic} subject-specific features (Section \ref{sec:motion_transfer}) from sequences of poses. 
\subsection{Static Pose Space Disentanglement}
\label{sec:motion_descriptor}
The regressor $R$ proposed in Section \ref{sec:regressor} relates body motion and body shape to soft-tissue deformations.
To represent body motion, a standard parameterization used across many human models \cite{feng2015reshaping,anguelov2005scape,loper2014mosh,loper_SIGAsia2015} is the joint angles of the kinematic skeleton, $\theta$.
However, our key observation is that this pose representation is entangled with shape- and subject-specific information that hinders the learning of a pose-dependent regressor.
Additionally, Hahn \textit{\etal} \cite{hahn2014subspace} also found that using joint angles to represent pose leads to a high-dimensional space with redundancies, which makes the learning task harder and prone to overfitting.
We hypothesize that existing data-driven parametric human models are less sensitive to this entanglement and overparameterization because they learn simpler deformations with much more data. In contrast, we model soft-tissue with a limited dataset of 4D scans, which requires a well disentangled and discriminative space to avoid overfitting tissue deformation and pose.
Importantly, notice that removing these features manually is not feasible, not only because of the required time, but also because these features are not always apparent to a human observer.

We therefore propose a novel and effective approach to \emph{deshape} the pose coefficients, \textit{i.e.}, to disentangle subject-specific anatomical features into a normalized and low-dimensional pose space $\bar{\theta}$:  
\begin{equation}
\bar{\theta} = E_\text{pose}(\theta).
\end{equation}
We find  $E_\text{pose}(\theta) \in \mathbb{R}^{|\bar{\theta|}}$ by training a multi-modal encoder-decoder architecture, shown in Figure~\ref{fig:pose_descriptor}. In particular, having a mesh scan $\mathbf{S}$ and its corresponding pose $\theta$ and shape $\beta$ parameters (found by solving Equation \ref{eq:fitting}), we simultaneously train two encoders and one decoder minimizing the loss
\begin{multline}
	\mathcal{L} =\norm{M(\theta,\mathbf{0}) - D_\text{mesh}(E_\text{mesh}(M(\theta,\mathbf{0}))}_2 \\
	 +\norm{M(\theta,\mathbf{0}) - D_\text{mesh}(E_\text{pose}(\theta)}_2,
\end{multline}
where $M(\theta,\mathbf{0}))$ are the surface vertices of a skinned mesh in pose $\theta$ and mean shape (\textit{\ie}, vector of shape coefficients is zero).
The intuition behind this multi-modal autoencoder is the following: the encoder $E_{\text{mesh}}$ takes as input \textit{skinned vertices} to enforce the similarity of large deformations (\textit{\eg}, lifting arms, where many vertices move) in the autoencoder loss. 
By using a significantly small latent space, we are able to simultaneously train it with
the encoder $E_{\text{pose}}$ such that the latter learns to remove undesired local pose articulations (and keep global deformations) directly in the pose vector $\theta$.
In contrast, notice that without the loss term that uses $E_{\text{mesh}}$ we would not be able to distinguish between large and small deformations, because in the pose parameterization space of $\theta$ all parameters (\textit{\ie}, degrees of freedom) contribute equally.

\begin{figure}
	\centering
	\includegraphics[width=0.9\columnwidth]{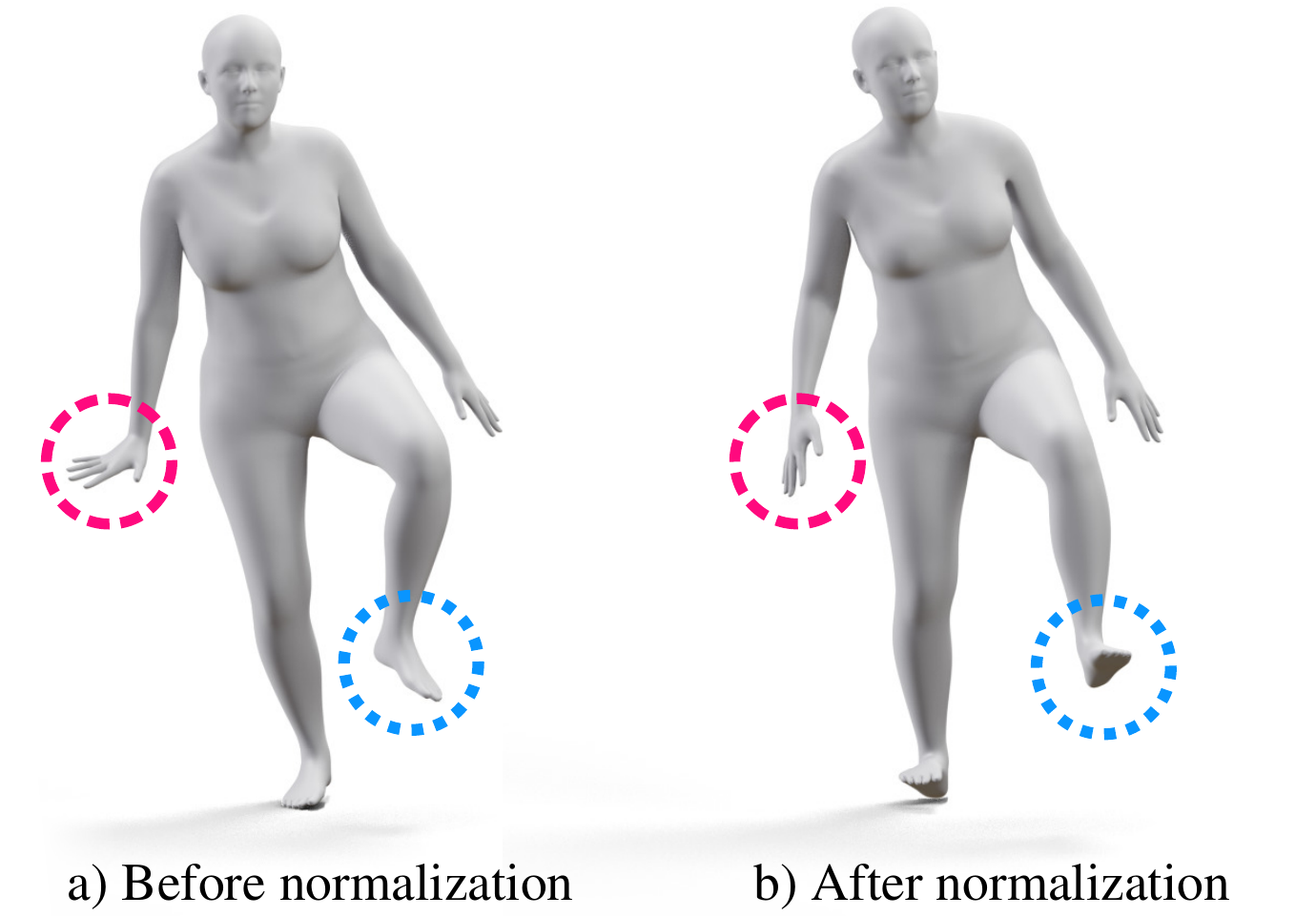}
	\caption{Result after static pose disentanglement. Our approach effectively removes subject- and shape-dependent features, while retaining the main characteristics of the input pose. See supplementary material for a visualisation of the pose disentanglement across a sequence.}
	\label{fig:pose_normalization}
\end{figure}

The effect of the encoder $E_{\text{pose}}$ is depicted in Figure \ref{fig:pose_normalization}, where subject- and shape-specific features are effectively removed, producing a \textit{normalized} pose.
In other words, we are disentangling features originally present in the pose descriptor $\theta$ (\textit{\eg}, wrist articulation) that are related to that particular subject or shape, but we are keeping the overall pose (\textit{\eg}, raising left leg).

We found 10 to be an appropriate size of the latent space for a trade-off between capturing subtle motions and removing subject-specific features.

\subsection{Avoiding Dynamic Pose Entanglement}
\label{sec:motion_transfer}
The novel pose representation $\bar{\theta}$ introduced earlier effectively disentangles \textit{static} subject-specific features from the naive pose representation $\theta$, however, our motion descriptor $\gamma$ also takes temporal information (velocities and accelerations) into account.
We observe that such temporal information can encode \textit{dynamic} shape- and subject-specific features, causing an entanglement potentially making our regressor prone to overfitting soft-tissue deformations to subject-specific pose dynamics. 

We address this by extending our 4D dataset by transferring sequences (encoded using our motion descriptor) across the different subjects. In particular, given two sequences of two different subjects
\begin{eqnarray}
\mathcal{S}^{i}_{\text{A}} &=& \{\textbf{S}^{i}_{\text{A},t}(\theta^{i}_{t})\}_{t=0}^{{N}^{i}_{A}}
\\
\mathcal{S}^{j}_{\text{B}} &=& \{\textbf{S}^{j}_{\text{B},t}(\theta^{j}_{t})\}_{t=0}^{{N}^{j}_{B}}
\end{eqnarray}
where $\textbf{S}^{i}_{\text{A},t}(\theta^{i}_{t})$ is the mesh of the subject $\text{A}$ performing the sequence identity $i$ at time $t$,
we transfer the sequence of poses $\theta_t^i$ to a subject $\text{B}$ by training a subject-specific regressor $R_B$. This process generates a new sequence 
\begin{eqnarray}
\mathcal{S}^{i}_{\text{B}} &=& R_{\text{B}}(\gamma^{~i}_{\text{~A}}) = \{\textbf{S}^{i}_{\text{B},t}(\theta^{i}_{t})\}_{t=0}^{{N}^{i}_{\text{A}}}
\end{eqnarray}
with the shape identity of the subject $\text{B}$ performing the motion $\theta_t^i$ (notice, a motion originally performed by subject $\text{A}$).  
By transferring all motions across all characters, we are enriching our dataset in a way that effectively avoids overfitting soft-tissue deformations to subject and shape-specific dynamics (\textit{\ie}, style).

In Section \ref{seq:evaluation} we detail the number of sequences and frames that we transfer, and evaluate the impact of this strategy. Specifically, Figure 	\ref{fig:shape_generalization_quantitative} shows an ablation study on how the generalization capabilities of our method improve when applying the pose disentangling methods introduced in this section.

%% file: data.tex
\section{Datasets, Networks and Training}
In this section we provide details about the datasets, network architectures, and parameters to train our models. 

\subsection{Soft-tissue Autoencoder and Regressor}
\paragraph*{Data.} Our soft-tissue autoencoder and soft-tissue regressor (Section~\ref{sec:regressor}) are trained using the 4D sequences provided in the Dyna dataset~\cite{ponsmoll2015dyna}. 
This dataset contains highly detailed deformations of registered meshes of 5 female subjects performing a total of 52 dynamic sequences captured at 60fps (42 used for training, 6 for testing).
Notice that we do not use the Dyna provided meshes directly, but preprocess them to \textit{unpose} the meshes. To this end, we solve Equation \ref{eq:fitting} for each mesh, and subsequently apply Equation \ref{eq:displacement_gt} to find the ground truth displacements for all Dyna meshes.

Moreover, in addition to the motion transfer technique described in Section~\ref{sec:motion_transfer}, we further synthetically augment the dataset by mirroring all the sequences.

\paragraph*{Setup.}
We implement all networks in TensorFlow, including the  encoder-decoder architecture of $E_{\text{soft}}$ and $D_{\text{soft}}$, and the $R$ regressor.
We also leverage TensorFlow and its automatic differentiation capabilities to solve Equation \ref{eq:fitting}. In particular, we optimize 
 $\beta$ using the first frame of a sequence and then optimize $\theta$ while leaving $\beta$ constant.
 We use Adam optimizer with a learning rate of 1e-4 for the autoencoder and 1e-3 for the regressor.
 The autoencoder is trained during 1000 epochs (around 3 hours) with a batch size of 256, and
 a dropout rate of 0.1. 
 The regressor is trained during 100 epochs (around 25 minutes) with batch size of 10, and no dropout. The details of the architecture are show\revised{n} in Figure~\ref{fig:pipeline}.

\subsection{Pose Autoencoder}

\paragraph*{Data.} 
To train our pose autoencoder presented in Section~\ref{sec:motion_descriptor} we are not restricted to the data of 4D scans because we do not need dynamics. 
We therefore leverage the SURREAL dataset \cite{varol_CVPR2017}, which contains a vast amount of Motion Capture (MoCap) sequences, from different actors, parameterized by pose representation $\theta$. Our training data consists of 76094 poses from a total of 298 sequences and 56 different subjects, including the 5 subjects of the soft-tissue dataset (excluding the sequences used for testing the soft-tissue networks).

\paragraph*{Setup.} We use Adam optimizer with a learning rate of 1e-3, and a batch size of 256, during 20 epochs (20 min). The details of the architecture are show\revised{n} in Figure~\ref{fig:pose_descriptor}.

%% file: evaluation.tex
\section{Evaluation}
\label{seq:evaluation}
In this section we provide qualitative and quantitative evaluation of both the reconstruction accuracy of our soft-tissue deformation subspace, described in Section \ref{sec:soft_tissue_subspace}, and the regressor proposed in Section \ref{sec:regressor}.
\subsection{Soft-tissue Autoencoder Evaluation}
\begin{figure}
	\includegraphics[width=\columnwidth]{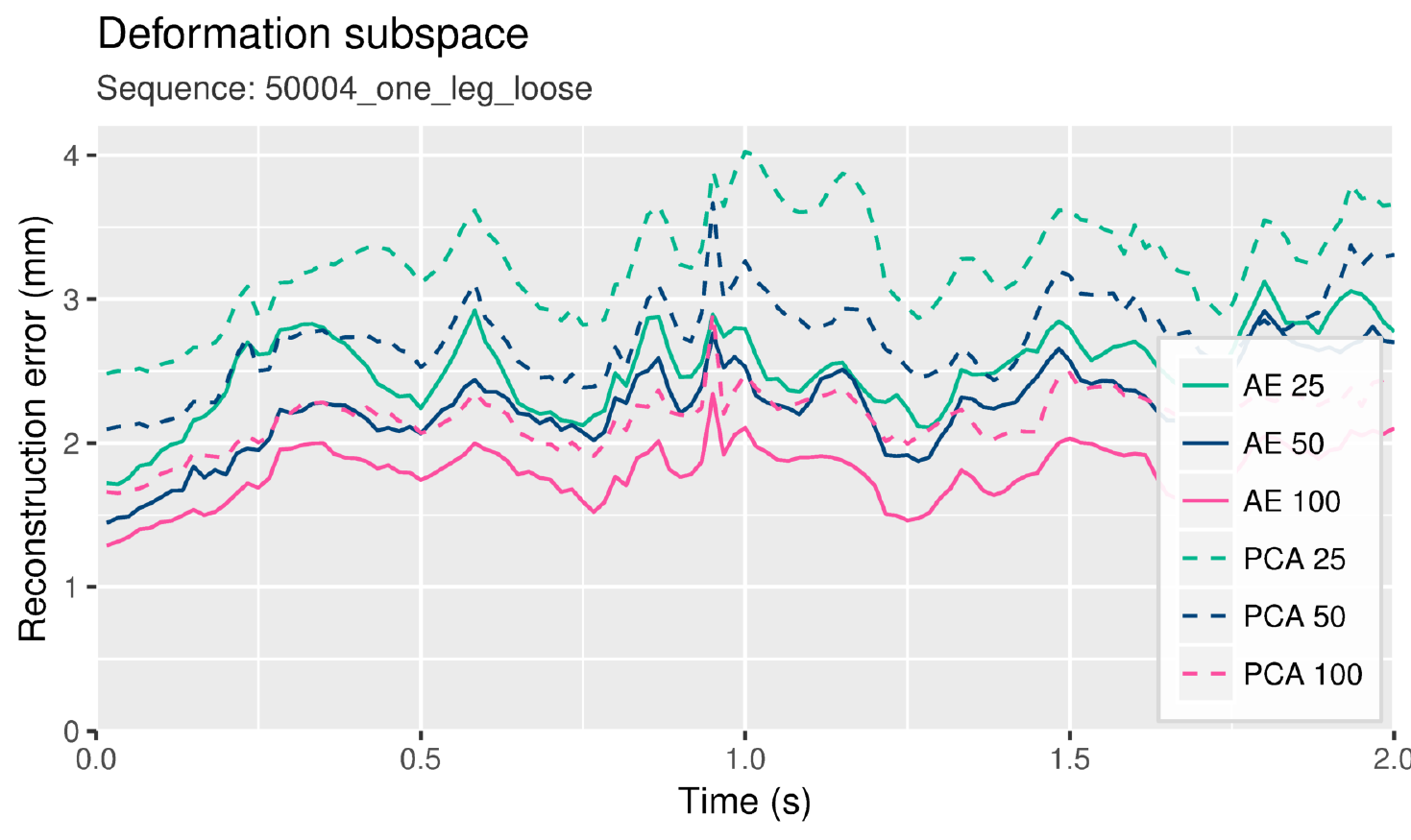}
	\caption{Soft-tissue autoencoder quantitative evaluation}
	\label{fig:autoencoder_quantitative_evaluation}
\end{figure}

\paragraph*{Quantitative Evaluation.} Figure \ref{fig:autoencoder_quantitative_evaluation} shows a quantitative evaluation of the reconstruction accuracy of the proposed nonlinear autoencoder (AE) for soft-tissue deformation, for a variety of subspace sizes.
We compare it with linear approaches based on PCA used in previous works \cite{loper_SIGAsia2015,ponsmoll2015dyna}, in a test sequence (\textit{\ie}, not used for training). \revised{Furthermore, Table \ref{table:autoencoder_full_dataset} shows the reconstruction error in the full test dataset.}
These results demonstrate that our autoencoder consistently outperforms the reconstruction accuracy of the subspaces used in previous methods.

\begin{table}[h] \centering

    \begin{tabular}{@{}lccc@{}}
    \toprule
        & 25D    & 50D    & 100D   \\ \midrule
    PCA & 3.82mm & 3.17mm & 2.38mm \\
    AE  & \textbf{3.02mm} & \textbf{2.58mm} & \textbf{2.09mm} \\ \bottomrule
    \end{tabular}

    \caption{\revised{Reconstruction error of our soft-tissue autoencoder and PCA evaluated in the full test dataset. The autoencoder (AE) performs better than the linear approach (PCA) in all tested subspace sizes.}}
    \label{table:autoencoder_full_dataset}
\end{table}

\begin{figure}
	\includegraphics[width=\columnwidth]{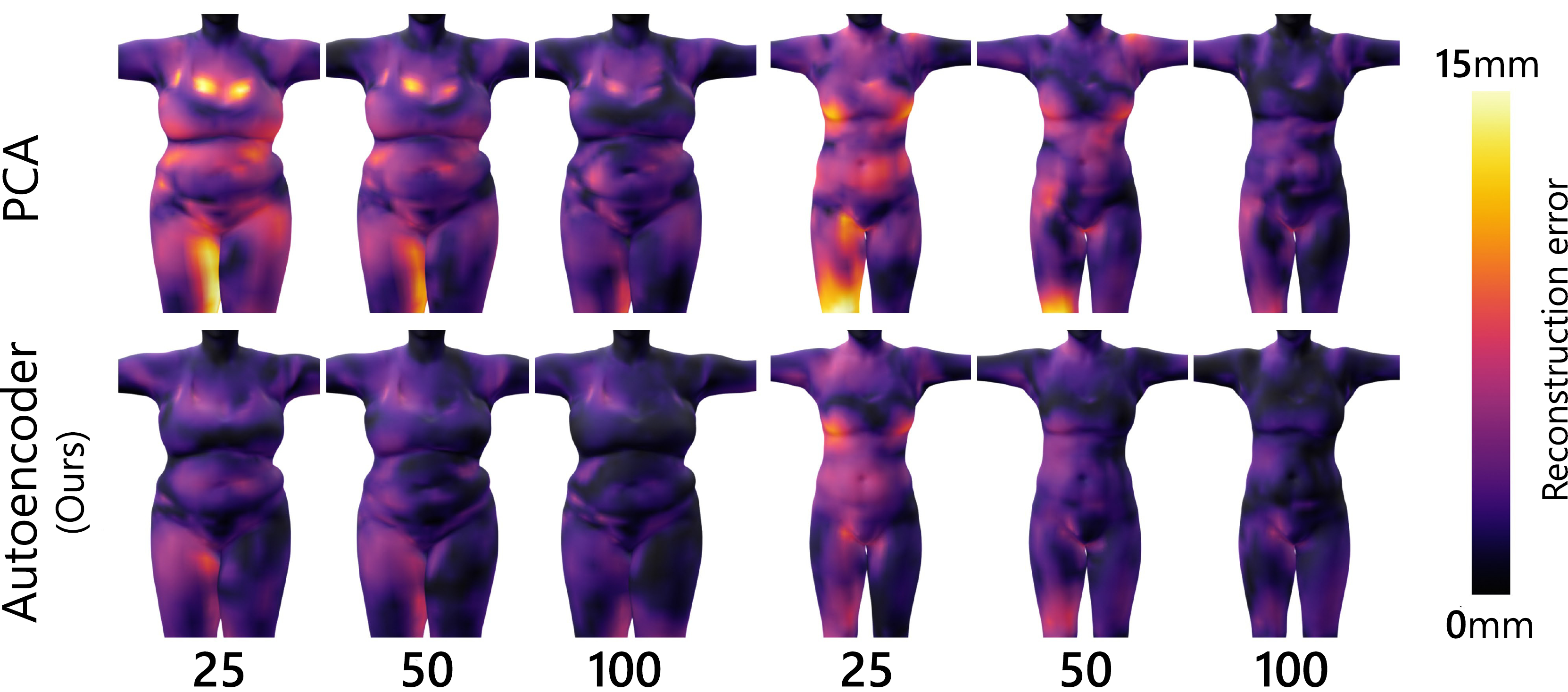}
	\caption{Reconstruction errors of our soft-tissue autoencoder and PCA, for two different body shapes. Notice that our subspace efficiently encodes soft-tissue displacements for parametric shapes, in contrast to previous works \cite{casas_PACMCGIT2018} that required an autoencoder per subject.}
	\label{fig:autoencoder_qualitative_evaluation}
\end{figure}

\paragraph*{Qualitative Evaluation.}
Figure \ref{fig:autoencoder_qualitative_evaluation} depicts a qualitative evaluation of the soft-tissue deformation autoencoder for a variety of subspace dimensions.
Importantly, we also show that the reconstruction accuracy is attained across different shapes. 
The realism of the autoencoder is better appreciated in the supplementary video, which includes highly dynamic sequences reconstructed with our approach.

\subsection{Soft-tissue Regressor Evaluation}
We follow a similar evaluation protocol as in Dyna \cite{ponsmoll2015dyna}, and evaluate the following scenarios to exhaustively test our method.
Additionally, we provide novel quantitative insights that demonstrate significantly better generalization capabilities of our regression approach with respect \revised{to} existing methods.
\paragraph*{Generalization to New Motions.}
In Figure
\ref{fig:regression_vs_groundtruth} and in the supplementary video we demonstrate the generalization capabilities of our method to unseen motions. In particular, at train time, we left out the sequence \texttt{one\_leg\_jump} of the Dyna dataset, and then use our regressor to predict soft-tissue displacements for this sequence, for the shape identity of the subject 50004.
Leaving ground truth data out at train time allows us to quantitatively evaluate this scenario. To this end, we also show a visualization of the magnitude of soft-tissue displacement for both ground truth $\Delta^{\text{GT}}$ and regressed $\Delta$ displacements, and conclude that the regressed values closely match the ground truth.

Additionally, in the supplementary video we show more test sequences of different subjects from the Dyna dataset animated with MoCap sequences from the CMU dataset \cite{varol_CVPR2017}. Notice that for these sequences there is no ground truth soft-tissue available (\textit{\ie}, actors were captured in a MoCap studio, only recording joint positions). 
Our animations show realistic and highly expressive soft-tissue dynamics that match the expected deformations for different body shapes.

\begin{figure}
	\includegraphics[width=\linewidth]{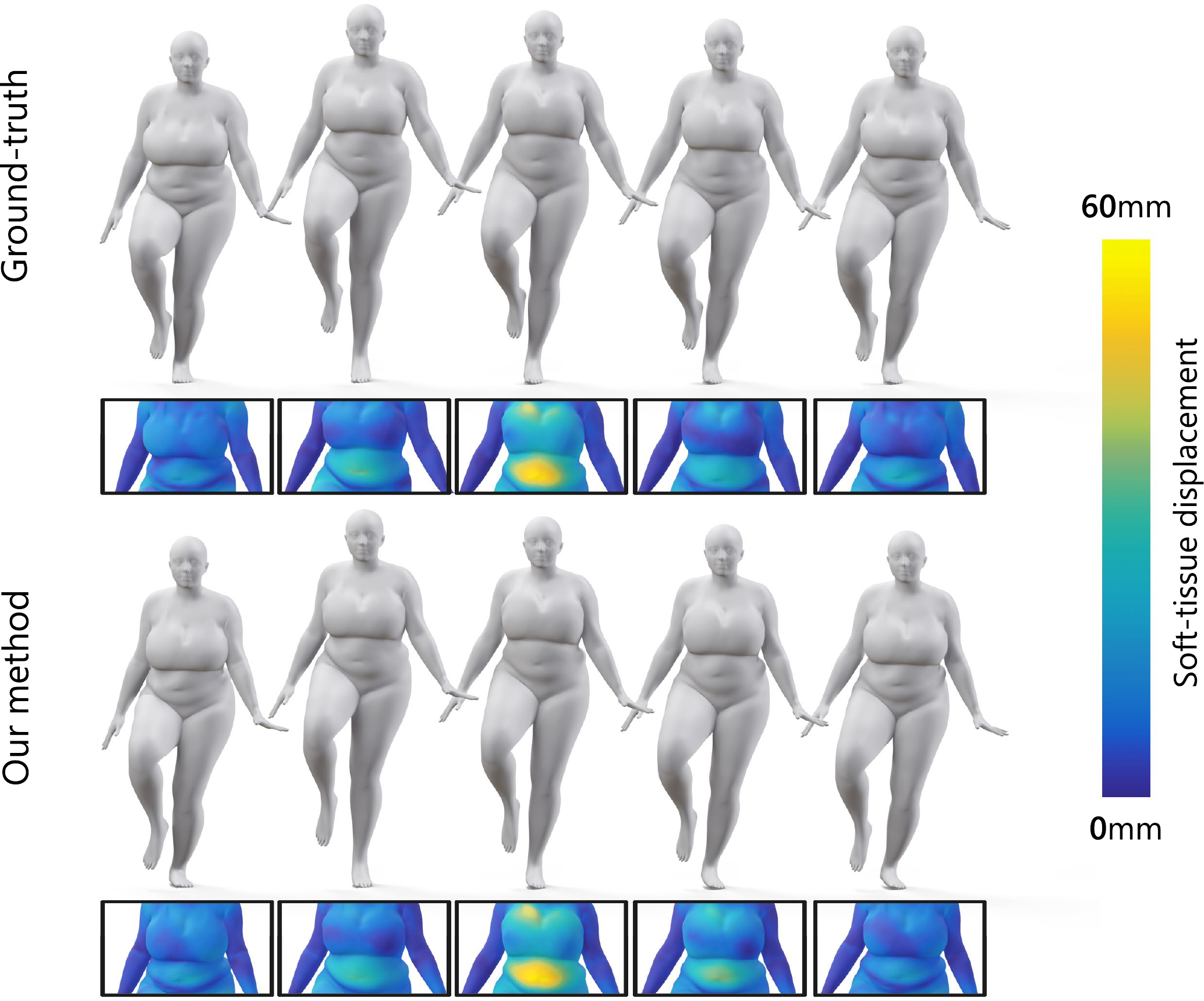}
	\centering
	\caption{Evaluation of generalization to new motions. The sequence \texttt{one\_leg\_jump} was left out at train time, and used only for testing, for subject 50004.  We show ground truth meshes and vertex displacements $\Delta^{\text{GT}}$ (top), and the regressed deformations $\Delta$ (bottom). Notice how the magnitude of the regressed displacement closely matches the ground truth.}
	\label{fig:regression_vs_groundtruth}
\end{figure}

\paragraph*{Generalization to New Subjects.}
We quantitatively evaluate the generalization capabilities to new subjects by looking at the magnitude of the predicted soft-tissue displacements for different body shapes.
Intuitively, subjects with larger body mass (\textit{i.e.}, more fat), which map to the smaller $\beta[1]$ parameters, should exhibit larger  soft-tissue velocities. In contrast, thin subjects, which maps to mostly positive values in $\beta[1]$, should exhibit much lower soft-tissue velocities due to the high rigidity of their body surface.
We exhaustively evaluate this metric in Figure \ref{fig:shape_generalization_quantitative}, where we show an ablation study comparing our full method, our method trained with each of the contributions alone, and Dyna.
\removed{As the original authors report in their paper \mbox{\cite{ponsmoll2015dyna}}, Dyna requires \textit{manually tweaked} scaling factors to output plausible surface dynamics for unseen subjects. If this manual exaggeration step is not done, Dyna generalization capabilities to new subjects are limited, producing surface dynamics with very similar deformations, as noted by the almost horizontal dark blue line in Figure \ref{fig:shape_generalization_quantitative}. }
\revised{Although Dyna \cite{ponsmoll2015dyna} produces different deformation modes for different subjects, the resulting motion is significantly attenuated.}
In contrast, our full model (in pink) regresses a higher dynamic range of deformations, output\revised{t}ing larger deformations for small values of $\beta[1]$ (\textit{\ie}, fat subjects), and small surface velocities for larger values of $\beta[1]$ (\textit{\ie}, thin subjects).
Importantly, we show that each contribution of our model (the static and dynamic pose disentangling methods introduced in Section \ref{sec:disentanglement}) contributes to our final results, and that all together produce the highest range of deformations.

In the supplementary video we further demonstrate our generalization capabilities. We also show an interactive demo where the user can change the shape parameters of an avatar in real\revised{-}time, and our method produces the corresponding and very compelling soft tissue deformation. 
\removed{To the best of our knowledge, this is the first time that a data-driven soft-tissue model is shown to generalize to parametric humans in real-time.} 

\begin{figure*}
	\includegraphics[width=\linewidth]{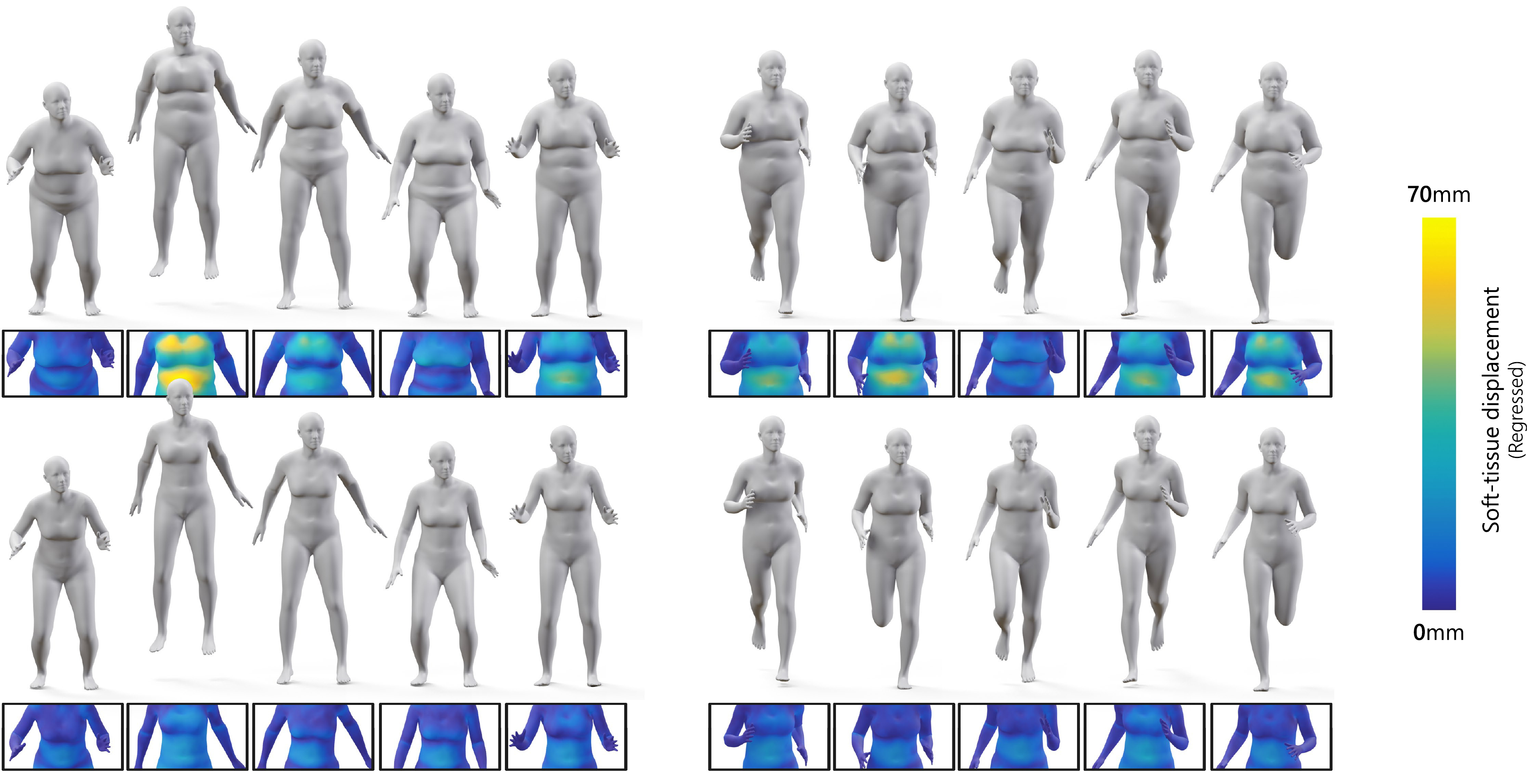}
	\caption{Sample frames of soft-tissue regression on two test sequences and two test subjects. Colormap depicts \revised{the} magnitude of the regressed deformation.
		Notice how our method successfully regresses larger  deformations on highly dynamic poses such as in the middle of a jump or when a \removed{food} \revised{foot} steps on the ground. See supplementary video for full animation and more examples.
	}
	\label{fig:regressor_qualiitative_meshes_colormaps}
\end{figure*}

{
	\fboxsep=0mm%
	\fboxrule=2pt%
\begin{figure}
	\includegraphics[width=\columnwidth]{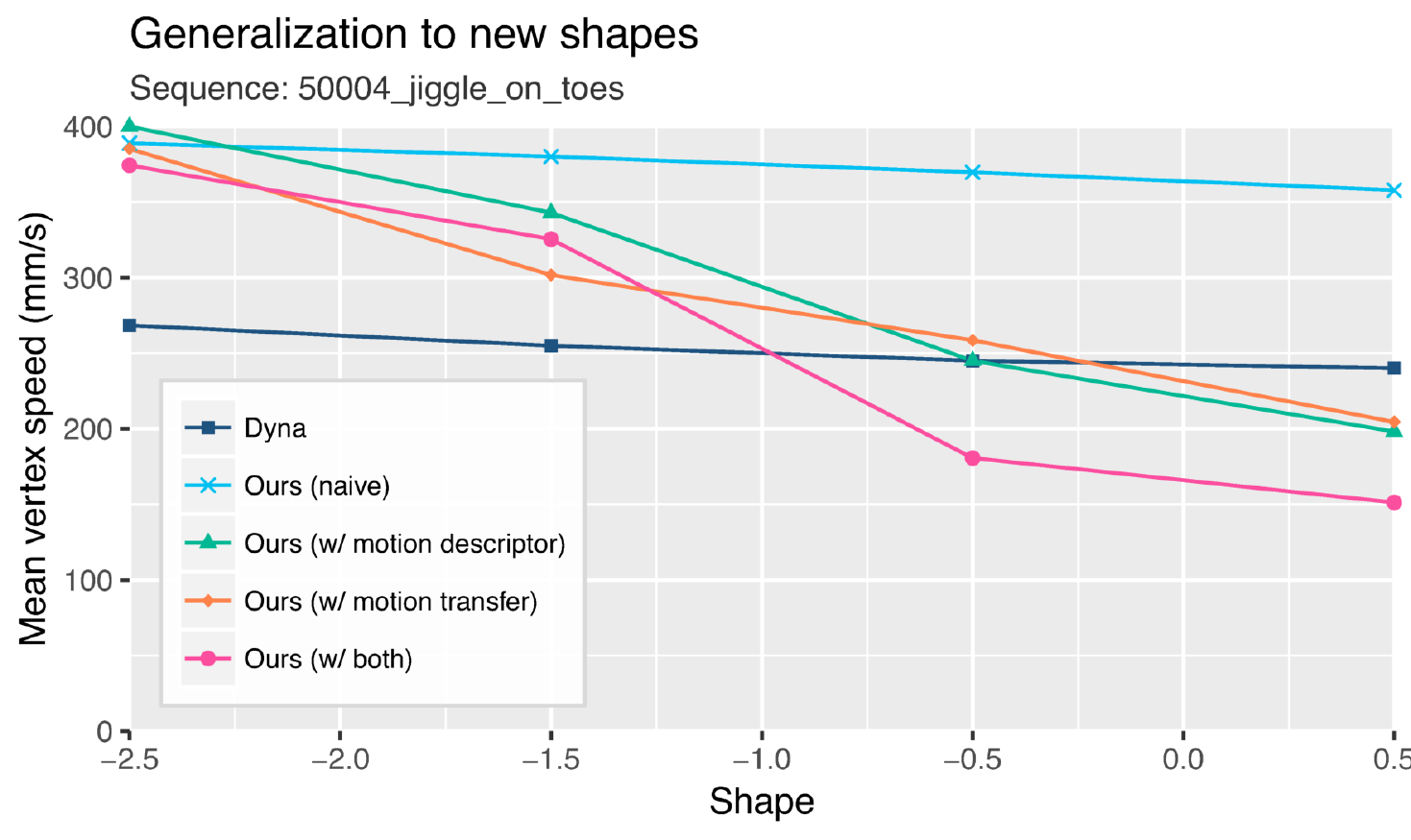}
	\includegraphics[width=\columnwidth]{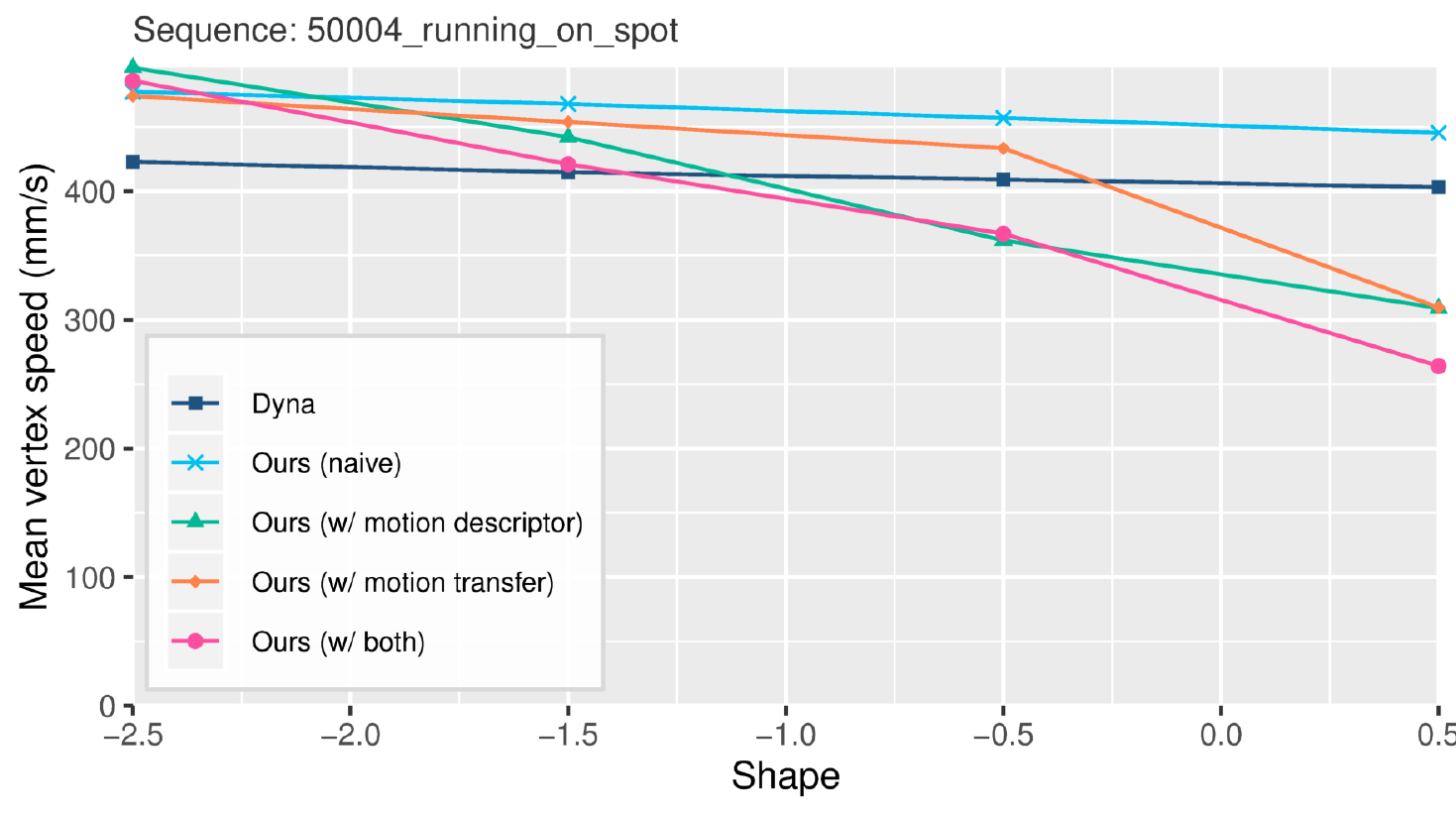}
	\caption{We quantitatively evaluate the generalization to new shapes of our regressor by looking at the mean vertex speed of the predicted soft-tissue offsets in unpose\revised{d} state \revised{in two test sequences}. Our model (pink) produces a higher range of dynamics, with large velocities for fat subjects (shape parameter -2.5) and small velocities for thin subjects (shape parameter 0.5).
		In contrast, previous works (Dyna, in dark blue) produce a much smaller range, resulting in limited generalization capabilities to new subjects.
Furthermore, here we also demonstrate that all components of our method contribute \removed{in} \revised{to} getting the best generalization capabilities.}
	\label{fig:shape_generalization_quantitative}
\end{figure}
}
\paragraph*{Generalization to New Motion and New Subject.} 
We finally demonstrate the capabilities of our model to regress soft-tissue deformations for new body shapes and motions.
To this end, we use MoCap data from SURREAL \revised{and AMASS} datasets \cite{varol_CVPR2017, mahmood2019amass} and arbitrary body shape parameters.
Figure \ref{fig:regressor_qualiitative_meshes_colormaps} shows sample frames of  sequences \texttt{01\_01} and \texttt{09\_10}
 for two different shapes. Colormaps on 3D meshes depict per-vertex magnitude regressed offsets to reproduce soft-tissue dynamics.
 As expected, frames with more dynamics exhibit larger deformations. Please see the supplementary video for more details.

\subsection{Runtime performance}
We have implemented our method on a regular desktop PC equip\revised{p}ed with an AMD Ryzen 7 2700 CPU, a Nvidia GTX 1080 GPU, and 32GB of RAM. \revised{After training the model, we use TensorRT \cite{tensorRT} to optimize the neural networks for faster inference at runtime}. On average, a forward pass of the \removed{full runtime} \revised{optimized} model takes \removed{6.5ms} \revised{4.8ms}.  This cost is distributed across the components of the model as follows: \removed{0.8ms} \revised{0.6ms} the pose encoder, \removed{2.6ms} \revised{1.9ms} the soft-tissue regressor and \removed{3.1ms} \revised{2.3ms} the soft-tissue decoder.

%% file: conclusions.tex
\section{Conclusions}
We have presented SoftSMPL, a data-driven method to model soft-tissue deformations of human bodies. 
Our method combines a novel motion descriptor and a recurrent regressor to generate per-vertex 3D displacements that reproduce highly expressive soft-tissue deformations.
We have demonstrated that the generalization capabilities of our regressor to new shapes and motions significantly outperform existing methods.
Key to our approach is to realize that traditional body pose representations rely on a\revised{n} entangled space that contains static and dynamic subject-specific features. 
By proposing a new disentangled motion descriptor, and a novel subspace and regressor, we are able to model soft-tissue deformations as a function of body shape and pose with unprecedented detail.

Despite the significant step forward towards modeling soft-tissue dynamics from data, our method suffers for the following limitations. With the current 4D datasets available, which contain very few subjects and motions, it is not feasible to learn a model for a high-dimensional shape space. Furthermore, subtle motions that introduce large deformations are also very difficult to reproduce.
Finally, as in most data-driven methods, our model cannot interact with external objects \revised{and does not support different topologies. Physics-based models can handle arbitrary meshes and react to external forces \cite{kim2017datadrivenphysics, kadlecek2016anatomical, komaritzan2018projective}, but they come at the expense of significantly higher computational cost.} 

\revised{Our approach to static pose disentanglement depends on compression, which is not always reliable and requires choosing an appropriate size for the pose space. Since the dataset contains several subjects performing similar motions, future works could make use of this information to find more robust ways to disentangle pose from static subject features.}

\revised{\paragraph*{Acknowledgments.} We would like to thank Rosa M. S\'anchez-Banderas and H\'ector Barreiro for their help in editing the supplementary video. Igor Santesteban was supported by the Predoctoral Training Programme of the Department of Education of the Basque Government (PRE\_2019\_2\_0104), and Elena Garces was supported by a Juan de la Cierva - Formaci\'{o}n Fellowship (FJCI-2017-32686). The work was also funded in part by the European Research Council (ERC Consolidator Grant no. 772738 TouchDesign) and Spanish Ministry of Science (RTI2018-098694-B-I00 \mbox{VizLearning}).}